\newif\iffinal
\finaltrue
\documentclass{article}
\usepackage{spconf,amsmath,epsfig}

\usepackage{hyperref}
\usepackage{url}
\usepackage[capitalise]{cleveref}
\usepackage{subcaption}
\usepackage{booktabs}
\usepackage[inline]{enumitem}
\usepackage{multirow}
\usepackage{paralist}
\usepackage{balance}

\title{Training general representations for remote sensing\\ using in-domain knowledge}
\iffinal
\name{Maxim Neumann, Andr\'{e} Susano Pinto, Xiaohua Zhai, and Neil Houlsby}
\address{Google Research, Zurich, Switzerland}
\else
\name{Names}
\address{Draft}
\fi

\newcommand{\imagenet}{ImageNet}
\newcommand{\bigearthnet}{BigEarthNet}
\newcommand{\eurosat}{EuroSAT}
\newcommand{\resisc}{RESISC--45}
\newcommand{\sosat}{So2Sat}
\newcommand{\ucmerced}{UC~Merced}
\newcommand{\ben}{\bigearthnet}
\newcommand{\eur}{\eurosat}
\newcommand{\res}{\resisc}
\newcommand{\sos}{\sosat}
\newcommand{\ucm}{\ucmerced}

\begin{document}
%
\maketitle

\begin{abstract}
Automatically finding good and general remote sensing representations allows to perform transfer learning on a wide range of applications -- improving the accuracy and reducing the required number of training samples. This paper investigates development of generic remote sensing representations, and explores which characteristics are important for a dataset to be a good source for representation learning. For this analysis, five diverse remote sensing datasets are selected and used for both, disjoint upstream representation learning and downstream model training and evaluation. A common evaluation protocol is used to establish baselines for these datasets that achieve state-of-the-art performance. As the results indicate, especially with a low number of available training samples a significant performance enhancement can be observed when including additionally in-domain data in comparison to training models from scratch or fine-tuning only on ImageNet (up to 11\% and 40\%, respectively, at 100 training samples). All datasets and pretrained representation models are published online.
\end{abstract}
%
\vspace*{-0.7em}
\section{Introduction}
\vspace*{-0.7em}


The number of Earth observing satellites is constantly increasing, with currently over 700 satellites monitoring many aspects of the Earth's surface and atmosphere from space, generating terabytes of imagery data every day that only automated machine learning systems will be able to process to retrieve all information of interest. At the same time, the ground truth label data -- as needed for good model training and calibration --  is costly to acquire, usually requiring extensive campaign preparation, people and equipment transportation, and in-field gathering of the characteristics under question.

Transfer learning is an approach that enables to pre-train \textit{upstream} a representation model on a large dataset, and to apply the learned knowledge \textit{downstream} to another related problem, for instance via fine-tuning on a specific target dataset, reducing significantly the number of required training samples. Often, ImageNet fine-tuning is used for knowledge transfer, but many other approaches exist \cite{zhai2019:vtab}.

In this paper we explore representation learning for remote sensing, and in particular in how much \textit{in-domain} knowledge from related datasets could help in representation learning. We look into what kind of data characteristics are important for good representation learning, and how the performance behaves at variable (especially smaller) downstream training sizes.

Recently, new large-scale remote sensing datasets have been generated, eg.\ \cite{zhu2018:so2sat, sumbul2019:bigearthnet, schmitt2019:sen12ms} that could be used for representation learning. However, a consistent evaluation framework is still missing and the performance is usually reported on non-standard splits and with varying metrics, making reproduction and quick research iteration difficult. To address this, we identified five representative and diverse remote sensing datasets and process them in a standardized form.
In summary, the main contributions of this work are:
\begin{compactitem}
    \item Exploring in-domain supervised fine-tuning to train generic remote sensing representations.
    \item Establishing performance baselines for the \ben, \eur, \res, \sos, and \ucm\ datasets, achieving state-of-the-art for datasets where comparison with past results was possible.
    \item Publishing 5 existing remote sensing datasets in a standardized format\footnote{Published at \href{https://www.tensorflow.org/datasets}{\tt www.tensorflow.org/datasets}.}, the used training splits\footnote{Published at \href{https://github.com/google-research/google-research/tree/master/remote\_sensing\_representations}{\tt github.com/google-research}.}
    and the trained representations\footnote{Published at 
    \href{https://tfhub.dev/google/collections/remote_sensing/1}{\tt www.tfhub.dev}.} for easy reuse by the community \cite{github_remote_sensing_representations}.
\end{compactitem}

\vspace*{-0.7em}
\section{Datasets}
\vspace*{-0.7em}
\label{sec:datasets}

Five datasets were selected for the analysis prioritizing newer and larger datasets that are quite diverse from each other, address scene classification tasks, and include at least optical frequency bands.

\begin{table*}
    \centering
    \caption{Used remote sensing datasets characteristics.}
    \label{tab:dataset_overview}
    \begin{tabular}{lllrrrrl}
    \toprule
    Name & year & Source & Size & Classes & Image size & Resolution & Problem \\
    \midrule
    \ben\ \cite{sumbul2019:bigearthnet} & 2019 & Sentinel-2 & ~590k & 43 & 20x20 to 120x120 & 10--60 m  & multi-label \\
    \eur\ \cite{helber2019:eurosat} & 2019 & Sentinel-2 & 27k & 10 & 64x64 & 10 m & multi-class\\
    \res\ \cite{cheng2017:resisc45} & 2017 & aerial & 31.5k & 45 & 256x256 & 0.2--60+ m & multi-class\\
    \sos\ \cite{zhu2018:so2sat} & 2019 & Sentinel-1/2 & ~376k & 17 & 32x32 & 10 m & multi-class\\
    \ucm\ \cite{yang2010:ucmerced} & 2010 & aerial & 2.1k & 21 & 256x256 & 0.3 m & multi-class\\
    \bottomrule
    \end{tabular}
\end{table*}

Dataset characteristics are summarized in \cref{tab:dataset_overview}. Differences between datasets include a wide range of image sizes and pixel resolutions, as well as the number of classes, dataset sizes, target domains, and differences in visual and semantic inter- and intra-class variances. More details are provided in \cite{neumann2020:iclr}.

For reproducability and a common evaluation framework, standard \emph{train, validation}, and \emph{test} splits using the 60\%, 20\%, and 20\% fractions, respectively, were generated for all datasets except \sos. For the \sos\ dataset, the original validation split is separated into validation and test splits with the 25\% and 75\% fractions, respectively.

\vspace*{-0.7em}
\section{Remote Sensing Data Processing}
\vspace*{-0.7em}

The remote sensing domain is quite distinctive from natural image domain and requires special attention during pre-processing and model construction. Some characteristics are:
\begin{compactitem}
    \item Remote sensing input data usually comes at higher precision (16 or 32 bits).
    \item The number of channels is variable, depending on the satellite instrument.
    \item The range of values varies largely from dataset to dataset and between channels. The values distribution can be highly skewed.
    \item Many quantitative remote sensing applications rely on the absolute values of the pixels.
    \item The images acquired from space are usually rotation invariant. 
    \item Source data can be delivered at different product levels. 
    \item Lower resolution data can aggregate a lot of information about the illuminated surface in a single pixel.
    \item Image axes might be non-standard, eg.\ representing range and azimuth dimensions.
\end{compactitem}

\vspace*{-0.7em}
\section{Approaches and Experimental Setup}
\vspace*{-0.7em}

The main goal is to develop representations that can be used across a wide range of unseen remote sensing tasks. The training and evaluation protocol follows two main stages: (1) \emph{upstream} training of the representations model based on some out- or in-domain data, and (2) \emph{downstream} evaluation of the representations by transferring the trained representation features to the new downstream tasks. For the \textit{upstream} training the full datasets are used. The \textit{downstream} training is performed using a pre-specified number of samples to assess the generalization of the trained representations and does never include any data that was used for upstream training.

All experiments use the same ResNet50 V2 architecture \cite{he2016:identity} and configuration (stochastic gradient descent (SGD) optimizer with momentum of 0.9, trained on TPU with batch sizes of 512, step-based learning rate decay ($\times 0.1$) at 4 stages with linear warm-up). 
Data pre-processing and augmentation can have a significant impact on performance and we used approaches. To compensate for the varied number of classes and training samples, a small set of hyper-parameter sweeps is performed, similar to \cite{zhai2019:vtab}:
\begin{compactitem}
\item 2 learning rates: $\{0.1, 0.01\}$,
\item 2 weight decays: $\{0.01, 0.0001\}$,
\item 3 training schedules: \{2500 steps with 200 warm-up steps; 10000 steps with 500 warm-up steps; 90 epochs with 5 warm-up\},
\item 2 data pre-processing approaches:\\
$simple$: resize original image to $224 \times\ 224$ at both train and eval.\\
$extended$: at train resize to $256\times\ 256$, random crop $224\times\ 224$, then random rotation and/or flip; at evaluation, resize to $256\times\ 256$ and central crop of $224\times\ 224$.
\end{compactitem}

For multi-class problems performance is reported using the Top-1 global accuracy metric, which denotes the percentage of correctly labeled samples.
For multi-label problems, the mean average precision (mAP) metric is used, which denotes the mean over the average precision values (integral over the precision-recall curve) of the individual labels.

\begin{table*}
    \caption{Performance of trained In-Domain and ImageNet representations (rows) when using only 1000 training examples for downstream tasks (columns).}
    \label{tab:cross-in-domain}
    \centering
\begin{tabular}{lrrrrr}
\toprule
Source \textbackslash Target & \bigearthnet{} &    \eurosat{} &   \resisc{} &     \sosat{} &  \ucmerced{} \\
\midrule
\imagenet{}    &       25.10 &      96.84 &      84.89 &      53.69 &      99.02 \\
\bigearthnet{} &           - &      96.45 &      78.43 &      50.91 &  \bf 99.61 \\
\eurosat{}     &       27.10 &          - &      79.59 &      52.99 &      98.05 \\
\resisc{}    &   \bf 27.59 &  \bf 97.14 &          - &  \bf 54.43 &  \bf 99.61 \\
\sosat{}      &       26.30 &      96.30 &      77.70 &          - &      97.27 \\
\ucmerced{}   &       26.86 &      96.73 &  \bf 85.73 &      53.52 &          - \\
\bottomrule
\end{tabular}
\end{table*}

\vspace*{-0.7em}
\section{Experimental Results}
\vspace*{-0.7em}

\subsection{Comparing in-domain representations}
\label{training-in-domain-repr}
\vspace*{-0.5em}

To obtain in-domain representations, first we train models either \emph{from scratch} or by \emph{fine-tuning} ImageNet on each full \emph{in-domain} dataset. The best of these models are then used as in-domain representations to train models on other remote sensing tasks (excluding the one used to train the in-domain representation).

For an initial evaluation of the different in-domain representation source data, \cref{tab:cross-in-domain} shows a cross-table evaluating each trained in-domain and ImageNet representation on each of the downstream tasks. The representations were trained using full datasets upstream, while the downstream tasks used only 1000 training examples to better emphasize the differences. As can be seen in this case, the best results all come from fine-tuning the in-domain representations. 

Despite having 2 distinctive groups of high-resolution aerial (\res{}, \ucm{}) and medium-resolution satellite datasets (\ben, \eur\ and \sos), the representations trained on \res\ were able to outperform the others in all tasks (\ben\ representations tied for the \ucm\ dataset) and it was the only representation to consistently outperform ImageNet-based representations. That \res\ would perform so good on both aerial and satellite tasks was unexpected. The reason is likely related to the fact that \res\ is the only dataset that has images with various resolutions. 
Combined with the large number of classes that have high within class diversity and high between-class similarity it seems to be able to train good representations for a wide range of remote sensing tasks, despite not being a very big dataset.

Counter to the expectation that bigger datasets should train better representations, the two biggest datasets, \ben{} and \sos{}, didn't provide the best representations (except of \ben\ representations for \ucm). We hypothesize that this might be due to the weak labeling and the low training accuracy obtained in these datasets. It is possible that the full potential of these large-scale datasets was not yet fully utilized and other specialized self- or semi-supervised representation learning approaches could improve the performance.

\begin{table}
    \centering
    \caption{Best performance on the selected remote sensing datasets.}
    \label{tab:benchmarks}
    \begin{tabular}{lrr}
    \toprule
    Dataset & Reference Result & Our Result\\
    \midrule
    \bigearthnet & 69.93\%/77.1\% (P/R)\cite{sumbul2019:bigearthnet}  & \textit{75.36}\% (mAP) \\
    \eurosat & 98.57\% \cite{helber2019:eurosat}  & \textbf{99.20}\% \\
    \resisc &  90.36\% \cite{cheng2017:resisc45} & \textbf{96.83}\% \\
    \sosat & & \textit{63.25}\% \\
    \ucmerced &  99.41\% \cite{nogueira2016:cnn} & \textbf{99.61}\% \\
    \bottomrule
    \end{tabular}
\end{table}

\begin{table*}
    \centering
    \caption{Accuracy over different training methods and number of used training samples.}
    \label{tab:overall}
\setlength{\tabcolsep}{3pt}
\begin{tabular}{l|rrr|rrr|rrr|rrr|rrr}
\toprule
{} & \multicolumn{3}{c|}{\bigearthnet} & \multicolumn{3}{c|}{\eurosat} & \multicolumn{3}{c|}{\resisc} & \multicolumn{3}{c|}{\sosat} & \multicolumn{3}{c}{\ucmerced} \\ 
{} & 100 & 1k & Full & 100 & 1k & Full & 100 & 1k & Full & 100 & 1k & Full & 100 & 1k & Full \\
\midrule
From Scratch  &            14.5 &             21.4 &             72.4 &        63.9 &         91.7 &         98.5 &         21.4 &          56.1 &          95.6 &       33.9 &        47.0 &        62.1 &          50.8 &           91.2 &           95.7 \\
ImageNet &            17.8 &             25.1 &         \bf 75.4 &        87.3 &         96.8 &         99.1 &         44.9 &          84.9 &          96.6 &       44.9 &        53.7 &        63.1 &          79.9 &           99.0 &           99.2 \\
InDomain &        \bf 18.8 &         \bf 27.6 &             69.7 &    \bf 91.3 &     \bf 97.1 &     \bf 99.2 &     \bf 49.0 &      \bf 85.7 &      \bf 96.8 &   \bf 46.4 &    \bf 54.4 &    \bf 63.2 &      \bf 91.0 &       \bf 99.6 &       \bf 99.6 \\
\bottomrule
\end{tabular}
\end{table*}

\vspace*{-0.7em}
\subsection{Large-scale comparison}
\vspace*{-0.7em}

Having trained in-domain representations, we can now evaluate and compare the transfer quality of fine-tuning the best in-domain representations with fine-tuning ImageNet and training from scratch at various training data sizes. 

As shown in \cref{tab:overall}, fine-tuning from ImageNet is better than training from scratch. And in all but one case, fine-tuning from an in-domain representation for transfer is even better.

The only exception is the \ben\ dataset at its full size. It is expected that having a large dataset should reduce the need for pre-training, but the gap between in-domain and ImageNet pre-training is quite big. We don't have an explanation for this yet and this needs to be further investigated.

Overall, these results establish new baselines for these datasets (some state-of-the-art), as summarized in \cref{tab:benchmarks}. Note that some results are not comparable: \res\ has been previously evaluated only on 20\% of data, \sos\ has no public benchmarking result to our knowledge, and the only published result of \ben{} is based on a cleaner version of the dataset (after removing the noisy images containing clouds and snow) and only precision and recall metrics were reported.

\vspace*{-0.7em}
\subsection{Small numbers of training examples regime}
\vspace*{-0.7em}

\begin{figure*}[tb]
     \centering
     \begin{subfigure}[b]{0.2\textwidth}
         \centering
         \includegraphics[width=\textwidth]{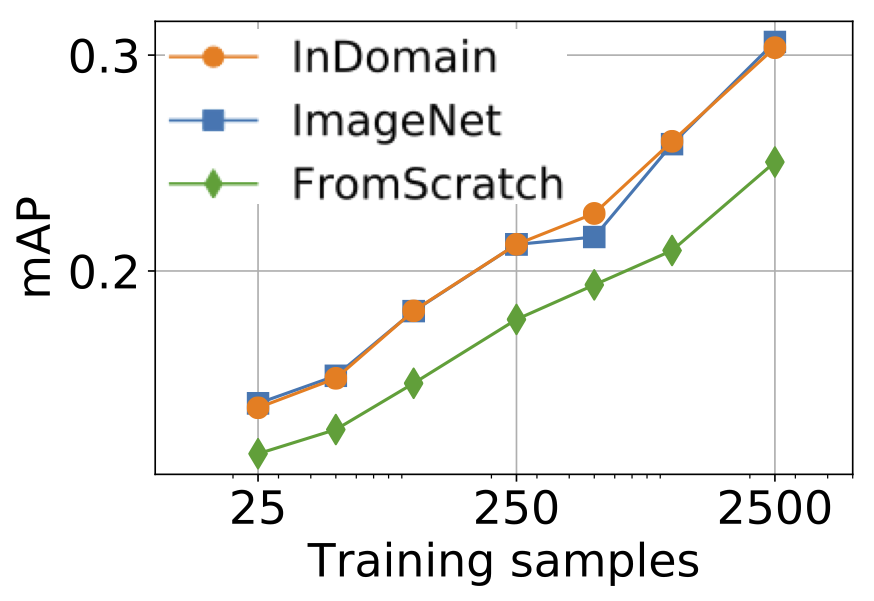}
         \caption{\bigearthnet{}}
         \label{fig:dense bigearthnet}
     \end{subfigure}%
     \begin{subfigure}[b]{0.2\textwidth}
         \centering
         \includegraphics[width=\textwidth]{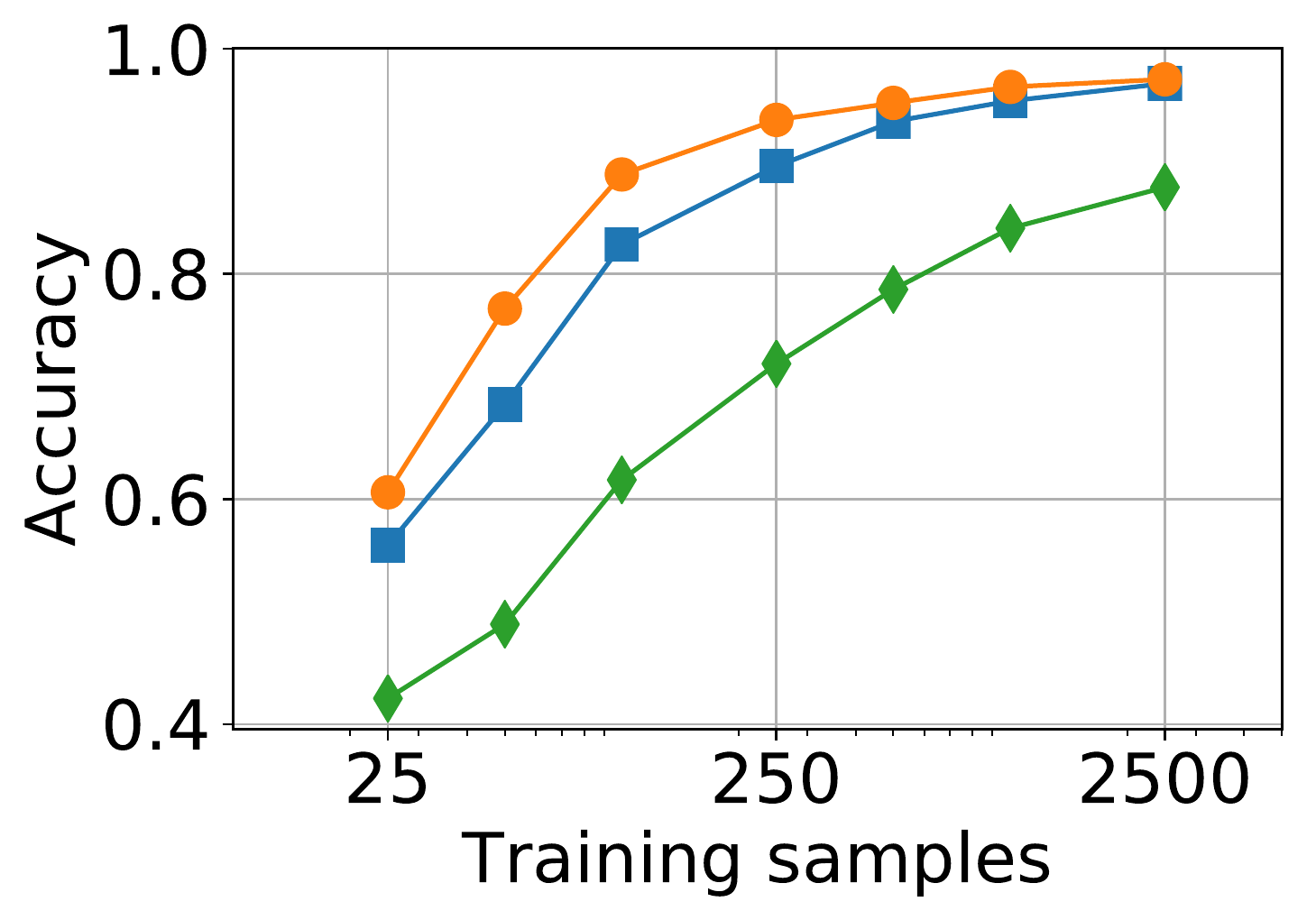}
         \caption{\eurosat{}}
         \label{fig:dense eurosat}
     \end{subfigure}%
     \begin{subfigure}[b]{0.2\textwidth}
         \centering
         \includegraphics[width=\textwidth]{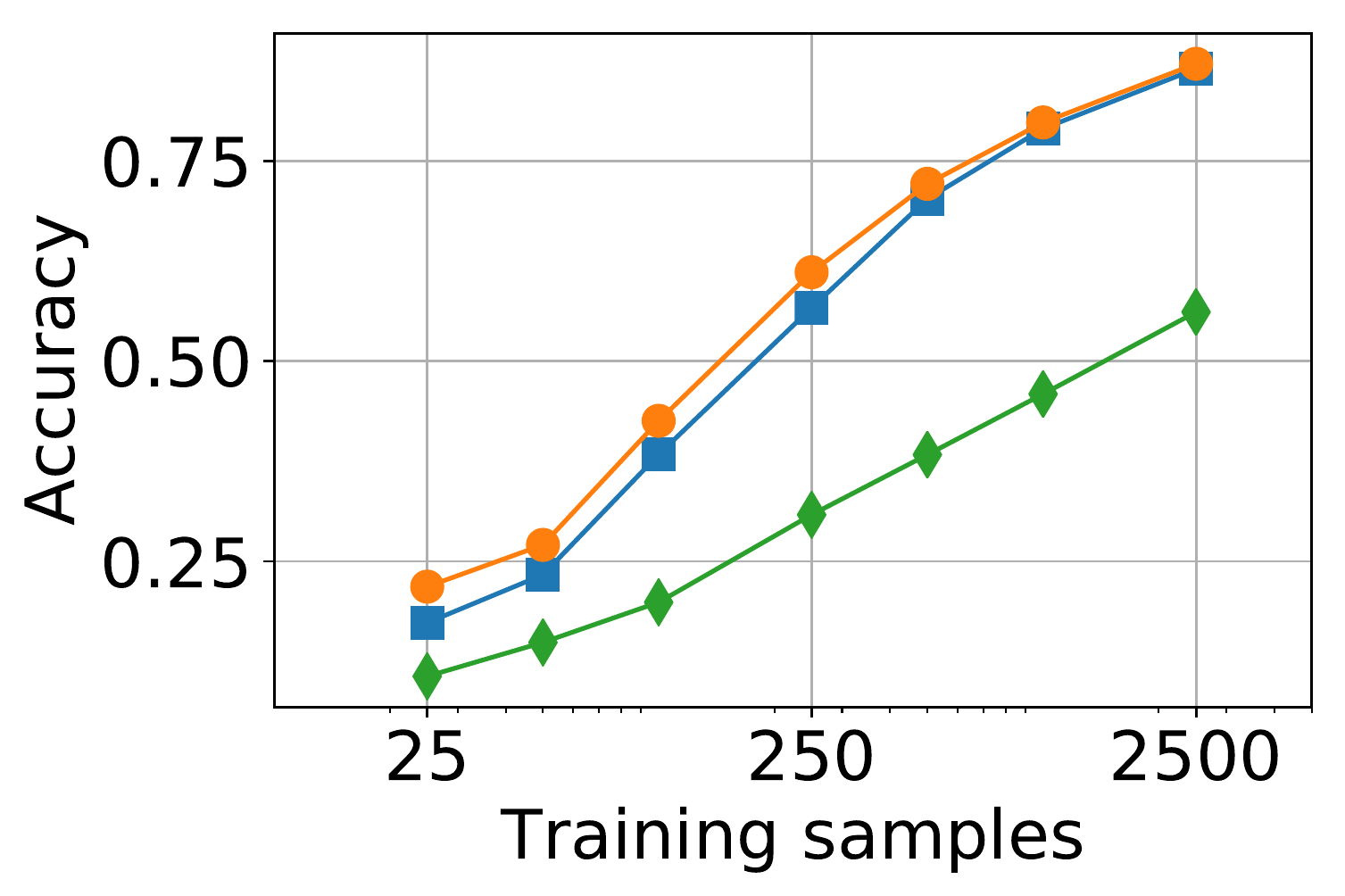}
         \caption{\resisc{}}
         \label{fig:dense resisc45}
     \end{subfigure}%
     \begin{subfigure}[b]{0.2\textwidth}
         \centering
         \includegraphics[width=\textwidth]{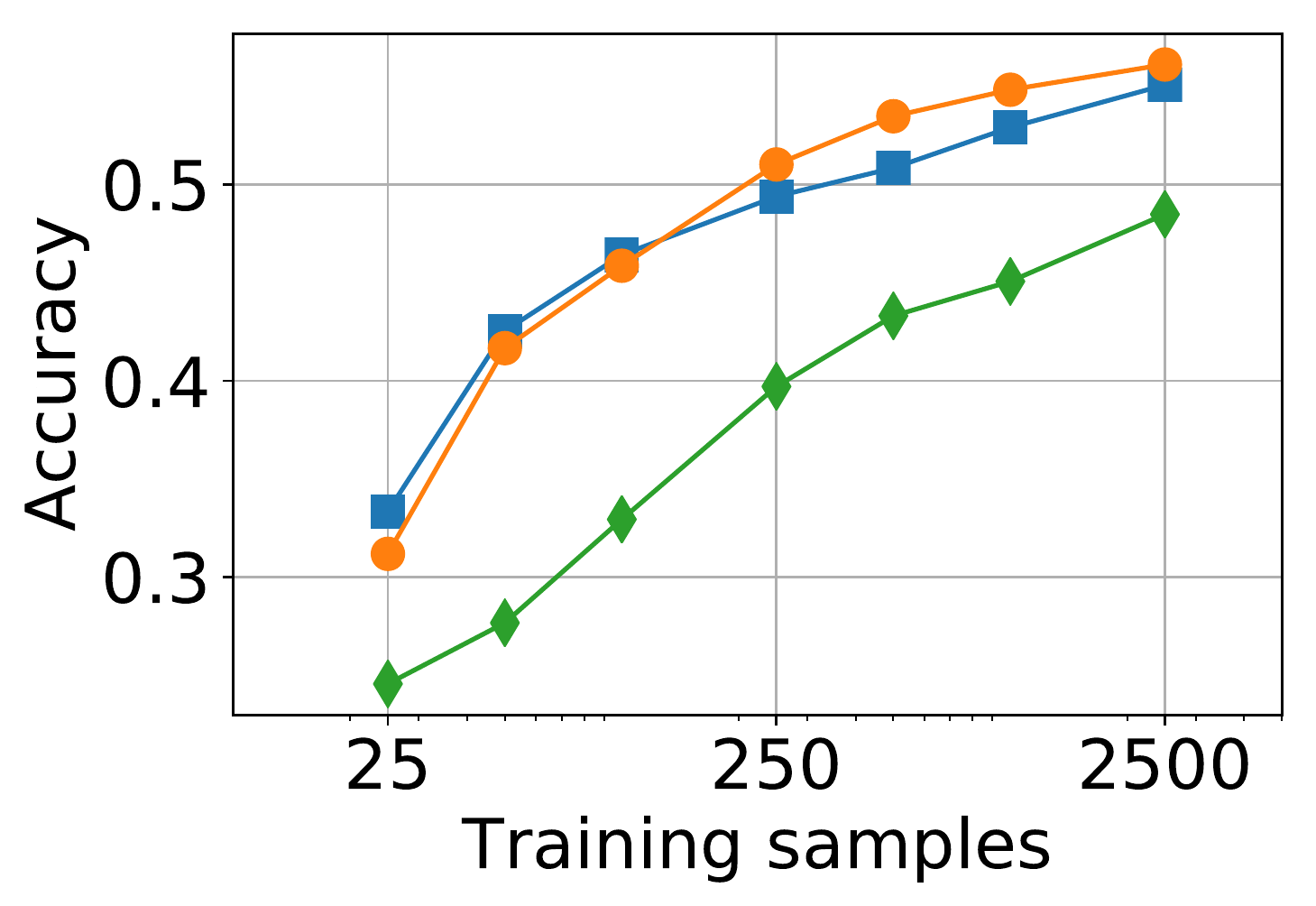}
         \caption{\sosat{}}
         \label{fig:dense so2sat}
     \end{subfigure}%
     \begin{subfigure}[b]{0.2\textwidth}
         \centering
         \includegraphics[width=\textwidth]{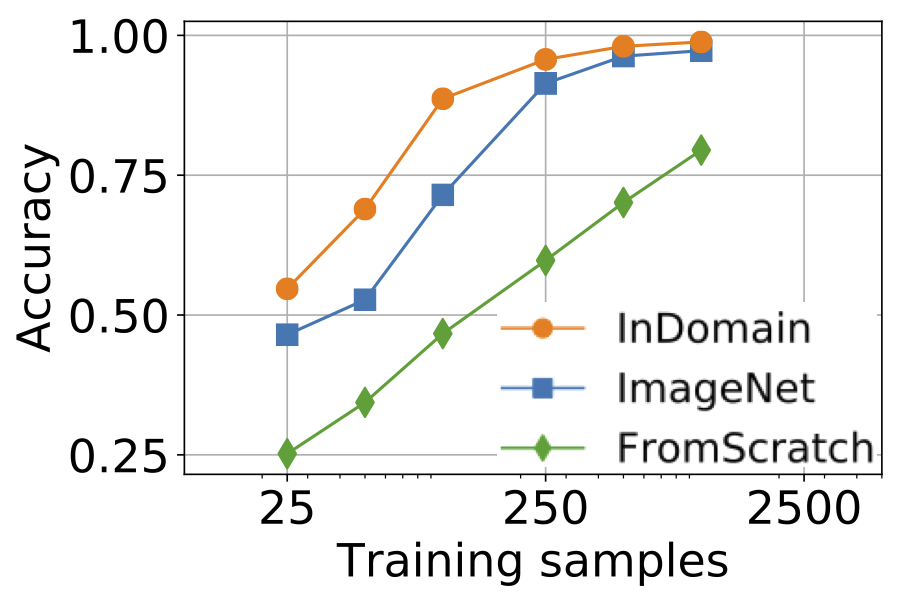}
         \caption{\ucmerced{}}
         \label{fig:dense uc_merced}
     \end{subfigure}

    \caption{Top-1 accuracy rate or mean average precision (mAP) over number of training examples.}
    \label{fig:dense}
\end{figure*}

To look closer into in-domain representation learning when only a small number of training examples is available, we performed transfer learning with small training sizes ranging from 25 to 2500 (samples were randomly drawn disregarding class distributions). We used a reduced set of hyper-parameters that might not deliver the most optimal performance, but still allows to observe the general trends.
As shown in \cref{fig:dense}, the improvement of using in-domain representations is clearly visible for the \eur, \res\ and \ucm\ datasets. These are the datasets with human-curated labels, which seem to be able to better profit from more specialized representations. The results are less conclusive for the \ben\ and \sos\ datasets that have more noisy labels.

\vspace*{-0.7em}
\subsection{Further Experiments}
\vspace*{-0.7em}

We performed more experiments using semi- and self-supervised approaches that are common in representation learning for natural images \cite{zhai2019:vtab}. However, since the results were worse, we excluded the other approaches from the analysis in this paper for brevity but will report them at the conference.

\vspace*{-0.7em}
\section{Conclusion}
\vspace*{-0.7em}

We present a common evaluation benchmark for remote sensing representation learning based on five diverse datasets. The results demonstrate the enhanced performance of \emph{in-domain} representations, especially for tasks with limited number of training samples, and achieve state-of-the-art performance on the full datasets. The five analyzed datasets and the best trained in-domain representations are published for easy reuse by the community in TFDS and TF-Hub.

As the experimental results indicate, having a multi-resolution dataset helps to train more generalizable representations. Other important factors seem to be label quality, number of classes, visual similarity across the classes and diversity within the classes. Surprisingly, we observed that representations trained on the large weakly-supervised datasets were not as successful as that of a smaller and more diverse human-curated dataset. However, some results were inconclusive and require more investigation. Understanding the main factors of a good remote sensing dataset for representation learning is a major challenge, solving which could improve performance  across a wide range of remote sensing tasks and applications. Other future directions include multi-task and multi-modality representation learning across a wide range of remote sensing data.

\balance
\small{
\bibliography{brain}
\bibliographystyle{IEEEbib}
}
\end{document}